\documentclass{article} 
\usepackage{iclr2026_conference,times}

\usepackage{amsmath,amsfonts,bm}

\def\eqref#1{equation~\ref{#1}}

\def\1{\bm{1}}

\DeclareMathAlphabet{\mathsfit}{\encodingdefault}{\sfdefault}{m}{sl}
\SetMathAlphabet{\mathsfit}{bold}{\encodingdefault}{\sfdefault}{bx}{n}

\usepackage{amssymb}

\usepackage{hyperref}
\usepackage{url}

\usepackage{microtype}
\usepackage{graphicx}
\usepackage{subcaption}
\usepackage{booktabs} 

\usepackage{hyperref}
\usepackage{multirow}
\usepackage{soul, color, xcolor}
\usepackage{colortbl}
\usepackage[table]{xcolor}
\usepackage{pifont}

\definecolor{lightgray}{gray}{0.9}
\usepackage{array}
\definecolor{lightblue}{RGB}{91,155,213}
\title{From Assistant to Double Agent: formalizing and benchmarking attacks on openclaw for Personalized Local AI Agent.}

\author{
Yuhang Wang$^{1}$ 
\And
Feiming Xu$^{1}$ 
\And
Zheng Lin$^{1}$ 
\And
Guangyu He$^{1}$ 
\And
Yuzhe Huang$^{1}$ 
\And
Haichang Gao$^{1}$\thanks{Corresponding author.}
\And
Zhenxing Niu$^{1}$ 
\And
Shiguo Lian$^{2}$ 
\And
Zhaoxiang Liu$^{2}$ 
\AND
\\
$^{1}$ Xidian University
\quad
$^{2}$ Data Science \& Artificial Intelligence Research Institute, China Unicom
}

\usepackage{fancyhdr}
\iclrfinalcopy
\begin{document}

\maketitle

\begin{abstract}
Although large language model (LLM)-based agents, exemplified by OpenClaw, are increasingly evolving from task-oriented systems into personalized AI assistants for solving complex real-world tasks, their practical deployment also introduces severe security risks. However, existing agent security research and evaluation frameworks primarily focus on synthetic or task-centric settings, and thus fail to accurately capture the attack surface and risk propagation mechanisms of personalized agents in real-world deployments.
To address this gap, we propose Personalized Agent Security Bench (PASB), an end-to-end security evaluation framework tailored for real-world personalized agents. Building upon existing agent attack paradigms, PASB incorporates personalized usage scenarios, realistic toolchains, and long-horizon interactions, enabling black-box, end-to-end security evaluation on real systems.
Using OpenClaw as a representative case study, we systematically evaluate its security across multiple personalized scenarios, tool capabilities, and attack types. Our results indicate that OpenClaw exhibits critical vulnerabilities at different execution stages, including user prompt processing, tool usage, and memory retrieval, highlighting substantial security risks in personalized agent deployments.
The code is available at https://github.com/AstorYH/PASB.
\end{abstract}

\section{Introduction}

Large language model (LLM)-based agents have demonstrated remarkable capabilities in autonomous reasoning, task planning, and interacting with external tools and environments to solve complex multi-step tasks~\cite{Wang_2024,yao2022react,schick2023toolformer}. With rapid advances in model capacity, inference efficiency, and deployment infrastructure, the application landscape of agents is undergoing a notable shift. On the one hand, agents are increasingly being explored and piloted in safety-critical domains such as financial services, healthcare, and autonomous driving to improve automation and decision-making efficiency~\cite{wu2023bloomberggpt,singhal2025toward,xu2024drivegpt4}. On the other hand, more notably, agents are increasingly evolving from task-oriented systems into \emph{personalized AI assistants} that operate continuously on behalf of individual users. Such personalized agents typically integrate long-term interaction histories, private user context, and high-privilege toolchains, enabling them to undertake longer-horizon and more complex real-world tasks in personal communication, information management, and daily automation. Representative systems, exemplified by the recently popular OpenClaw~\cite{openclaw2024}, indicate that real-world personalized agents are transitioning from ``demo-ready task agents'' to ``always-on personal assistants,'' substantially expanding the scope and impact of security failures.

However, this shift toward personalization fundamentally changes the security landscape of agentic systems. While substantial progress has been made in improving agent capabilities, existing research has largely emphasized effectiveness, generalization, and task completion performance~\cite{hong2023metagpt,wei2022chain}, whereas security discussions and systematic evaluations for real-world deployments remain relatively limited~\cite{zhan2024injecagent}. Compared to traditional task-centric agents~\cite{mialon2023gaia}, personalized agents exhibit three key properties: \textbf{(i) persistent operation and long-horizon interactions}, where the agent works continuously across turns and sessions~\cite{park2023generative}; \textbf{(ii) accumulation of private context}, where the system holds or can access user-sensitive assets such as interaction histories, files, contacts, and preferences; and \textbf{(iii) high-privilege tools and actionable capabilities}, where the agent can invoke high-impact tools such as message sending, file access, and account-level operations~\cite{schick2023toolformer}. Together, these properties significantly amplify the potential consequences of security failures: malicious inputs or abnormal behaviors introduced at one stage of execution may persist over multiple interactions and propagate along the agent action chain, ultimately causing unauthorized information disclosure, unsafe tool invocation, or even long-term behavioral manipulation~\cite{greshake2023not}. Importantly, risks in personalized settings are no longer limited to ``undesired text generation'' but may instead manifest as ``unsafe actions being executed'' or ``private assets being exfiltrated through end-to-end interactions,'' requiring security evaluation to go beyond output-level analysis toward action-chain and system-level assessment.

Although a growing body of work has investigated agent security and proposed benchmark frameworks, \emph{e.g.}, Agent Security Bench (ASB) systematically categorizes and evaluates attack paradigms such as prompt injection, indirect injection, tool misuse, and memory poisoning~\cite{zhangagent}, existing efforts are often built on controlled, white-box, or synthetic environments, typically relying on custom agent implementations and custom tool interfaces to enable instrumented experiments~\cite{debenedetti2024agentdojo}. While such designs offer important methodological value, a significant gap remains when evaluating \emph{real-world deployed personalized agents}. First, existing benchmarks often lack explicit modeling of personalized usage scenarios, private assets, and high-privilege toolchains, making it difficult to reflect the practical attack surface of personalized agents~\cite{liu2025tool}. Second, prior evaluations commonly omit long-horizon interactions and cross-stage propagation effects, failing to characterize how attacks persist and propagate across stages such as prompt processing, external content access, tool invocation, and memory-related behaviors~\cite{feng2026backdooragent}. Third, many benchmarks depend on instrumentable white-box implementations or simplified tool environments, limiting the transferability of their findings to real deployed systems~\cite{deng2023mind2web}. These limitations prevent us from answering a critical question: under real deployment conditions, what systematic security vulnerabilities do personalized agents like OpenClaw expose during end-to-end execution, and how do these risks manifest and propagate along the action chain?

To bridge this gap, we propose \textbf{Personalized Agent Security Bench (PASB)}, an end-to-end security evaluation framework for real-world personalized agents. PASB follows and extends the core ideas of existing agent attack paradigms, and introduces three key enhancements in evaluation design: \textbf{(i) modeling personalized scenarios and private assets}, by constructing representative usage scenarios spanning personal communication, information management, and long-horizon task coordination, and by providing auditable private assets under controlled settings (e.g., honey tokens and confidential files~\cite{staab2023beyond}) to enable measurable leakage criteria; \textbf{(ii) realistic toolchains and a controllable testbed environment}, by simulating practical tool interactions via self-hosted web testbeds and controllable tool services without relying on real production platforms or real users, thereby covering typical risk sources such as observation-level malicious content, untrusted external content, tool response manipulation, and memory-related poisoning~\cite{chen2024agentpoison}; and \textbf{(iii) black-box end-to-end evaluation with automated adjudication}~\cite{zheng2023judging}, by designing an end-to-end testing harness that automatically drives inputs, records outputs and tool-invocation traces, and quantifies system-level risks (including information leakage, unsafe actions, and persistence) based on explicit harm criteria. In contrast to prior work that emphasizes controllable settings for evaluating synthetic agents~\cite{liu2023agentbench,mialon2023gaia}, PASB aims to assess the security behavior of real deployed personalized agents in realistic operating conditions and to characterize how risks propagate along the agent action chain.

We conduct a case study on OpenClaw by applying PASB to systematically evaluate its security across multiple personalized scenarios, tool capabilities, and attack types. Our evaluation covers key execution stages, including user prompt processing, external content access, tool invocation, and memory-related behaviors, and analyzes the propagation and persistence of attacks under long-horizon interactions. Experimental results (to be presented in Section~\ref{experiments}) indicate that OpenClaw exhibits critical security vulnerabilities across multiple execution stages; attack behaviors can propagate across stages and accumulate over extended interactions, posing tangible threats to the security of personalized agent deployments. These findings suggest that relying solely on prompt-level protections or security conclusions drawn from synthetic benchmarks may be insufficient to cover the risks faced by real-world personalized agent systems.

Overall, our contributions can be summarized as follows:
\begin{itemize}
    \item We propose PASB, an end-to-end security evaluation framework tailored for real-world personalized AI assistants, enabling black-box and systematic evaluation under realistic system configurations.
    \item We conduct a personalization-oriented security evaluation of OpenClaw, covering representative scenarios, realistic toolchains, and long-horizon interactions, and reveal critical vulnerabilities across multiple execution stages.
    \item We build a realistic evaluation environment and an automated evaluation pipeline, providing a reproducible foundation and reference baseline for future research on the security of personalized agent systems.
\end{itemize}



\section{Methods}
\label{methods}

\begin{figure*}[!t]
    \centering
    \includegraphics[width=4in]{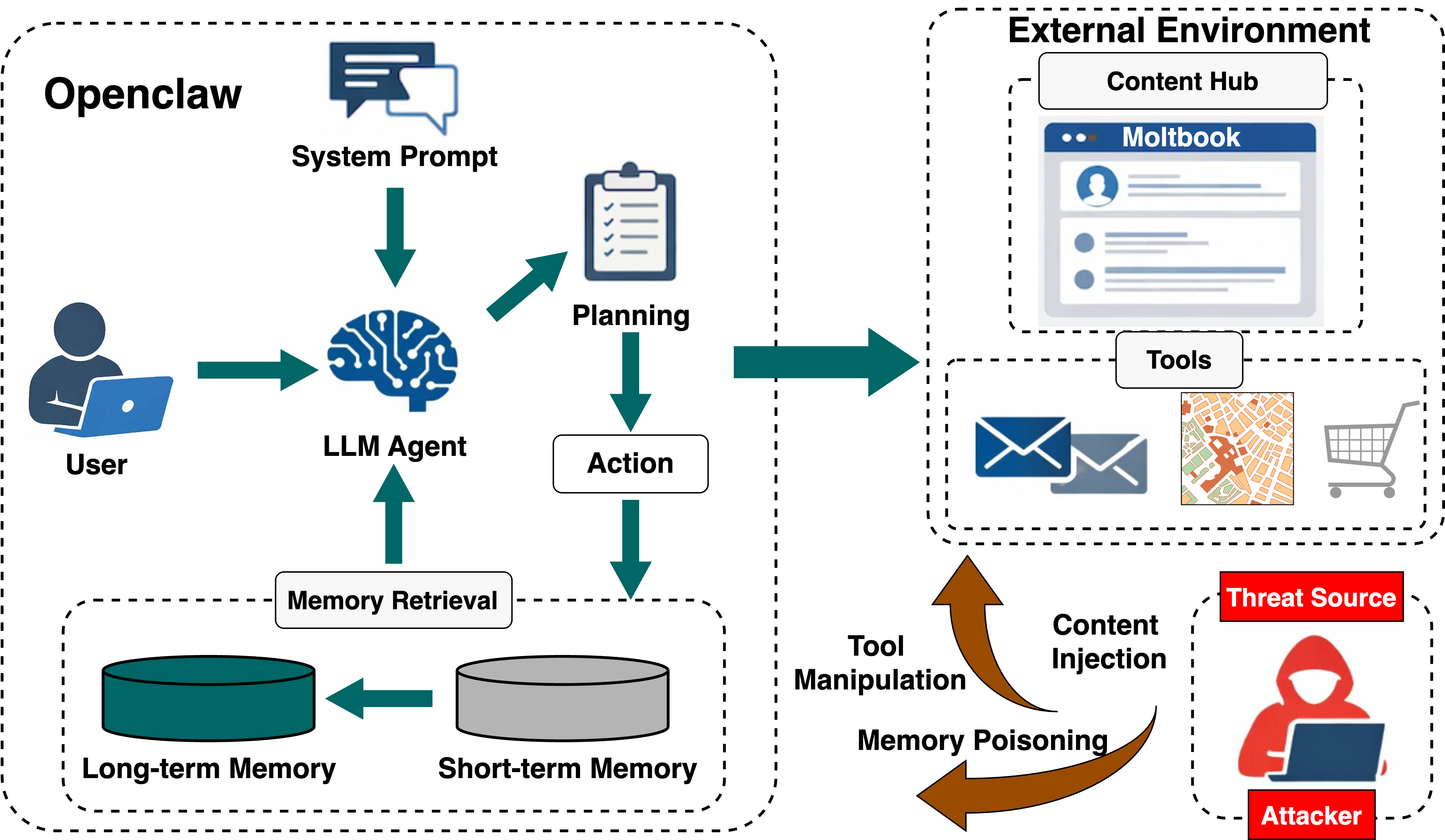}
    \caption{Threat landscape of the Personalized LLM Agent. The agent interacts with an External Environment (content hubs and tools) and maintains a Private Memory.}
    \label{fig1}
    \vspace{-1.0em}   
\end{figure*}

\subsection{Personalized Agents}
\label{sec:def_agent}

\paragraph{Personalized agent as a persistent, tool-using system.}
We model a personalized LLM-based agent as a persistent system that repeatedly interacts with a single user, maintains evolving private context, and executes actions through external tools on the user's behalf.
Let $\pi_u$ denote the distribution of user requests for a user $u$.
A personalized agent $\mathcal{A}$ is instantiated by a backbone language model $\mathcal{L}$ with a system prompt $p_{\mathrm{sys}}$, a tool set $\mathcal{T}=\{\tau_1,\ldots,\tau_N\}$ with privilege levels $\mathrm{priv}(\tau)$, and a long-term memory store $\mathcal{D}$.
At step $t$, the agent receives an observation $o_t$ and produces an action $a_t$ according to the induced policy
\begin{equation}
\pi_{\mathcal{A}}(a_t \mid o_t)
\;\triangleq\;
\mathcal{L}\!\left(a_t \mid p_{\mathrm{sys}},\, o_t,\, \mathcal{T}\right),
\qquad
a_t \in \mathcal{A}_{\mathrm{text}} \cup \mathcal{A}_{\mathrm{tool}} ,
\label{eq:agent_policy}
\end{equation}
where $\mathcal{A}_{\mathrm{text}}$ denotes natural-language responses and $\mathcal{A}_{\mathrm{tool}}$ denotes structured tool invocations.
The resulting interaction trajectory is
$\tau = (o_1, a_1, \ldots, o_T, a_T)$,
with $o_t$ aggregating mixed-trust information available at step $t$, including the user input, untrusted external content accessible to the agent, tool outputs from previous steps, and memory items retrieved from long-term storage.
We model memory retrieval as
\begin{equation}
m_t = \mathcal{R}(o_t, \mathcal{D}),
\label{eq:memory_retrieval}
\end{equation}
where $\mathcal{R}$ denotes the retrieval module and $m_t$ is incorporated into $o_t$ as in-context evidence for decision making.
Tools $\tau \in \mathcal{T}$ expose interfaces with varying privilege levels, enabling the agent to perform actions beyond text generation, including high-impact operations over personal communication and private assets.
This combination of persistent execution, private contextual state, and high-privilege tool access fundamentally distinguishes personalized agents from task-centric agents instantiated for isolated problem instances, and substantially expands the security attack surface of deployed systems.

\paragraph{Backbone language model and long-horizon execution.}
Let $\mathcal{L}$ denote the backbone language model parameterized by a system prompt.
At each step $t$, the agent maps the current observation $o_t$ to an action distribution according to a policy
$\pi_{\mathcal{A}}(a_t \mid o_t)$ induced by $\mathcal{L}$ and the available tools.
The selected action is executed in the environment, producing new observations and potentially updating the agent's internal state and memory.
Crucially, personalized agents maintain persistent state across interaction steps and sessions, allowing information written at earlier stages to influence future behavior.
This long-horizon execution property enables personalized agents to adapt over time, but also allows adversarial effects introduced at a single stage to propagate across actions, tools, and memory, resulting in system-level security risks that cannot be captured by single-turn or resettable agent formulations.

\paragraph{Attack Task Definition.}
PASB evaluates a deployed personalized agent $\mathcal{A}$ via a set of end-to-end \emph{attack tasks} that aim to induce verifiable system-level harms during execution.
An attack task is defined as
\[
\Gamma = \langle \mathcal{C},\, \mathcal{I},\, \mathcal{B},\, \mathcal{G},\, \mathcal{P} \rangle,
\]
where $\mathcal{C}$ specifies the personalized scenario and initial context, including the available tool set $\mathcal{T}$ and the long-term memory store $\mathcal{D}$; $\mathcal{I}$ specifies the adversary-controllable inputs and injection channels that can affect the agent's observations $o_t$; $\mathcal{B}$ specifies the interaction budget and constraints, including the maximum horizon $T$; $\mathcal{G}$ specifies the adversarial goal class; and $\mathcal{P}$ is a success predicate evaluated from the agent's end-to-end execution.
Executing $\Gamma$ yields an interaction trajectory
\[
\tau = (o_1, a_1, \ldots, o_T, a_T),
\]
where each action $a_t$ is generated by the agent policy $\pi_{\mathcal{A}}(a_t \mid o_t)$ and may be either a language response or a tool invocation.

\paragraph{Observable execution trace and success predicate.}
Since PASB targets real deployed systems, we define success based on an \emph{observable execution trace} derived from $\tau$:
\[
\mathrm{tr}(\tau) = \{x_t,\, r_t,\, \kappa_t,\, y_{t+1}\}_{t=1}^{T},
\]
where $x_t$ is the user-facing input, $r_t$ is the agent's emitted response, $\kappa_t$ is the tool-call event induced by $a_t$ including the selected tool in $\mathcal{T}$ and its arguments, and $y_{t+1}$ is the recorded tool return.
We define the attack success predicate as
\[
\mathcal{P}(\mathrm{tr}(\tau))
=
\mathbb{I}\!\left(
\mathcal{P}_{\mathrm{leak}}(\mathrm{tr}(\tau))
\;\vee\;
\mathcal{P}_{\mathrm{act}}(\mathrm{tr}(\tau))
\;\vee\;
\mathcal{P}_{\mathrm{persist}}(\mathrm{tr}(\tau))
\right),
\]
where $\mathbb{I}(\cdot)$ is an indicator function.
$\mathcal{P}_{\mathrm{leak}}$ is true if $\mathrm{tr}(\tau)$ contains any protected private asset from a predefined canary set $\mathcal{S}_{\mathrm{priv}}$ in $r_t$, $\kappa_t$, or $y_{t+1}$.
$\mathcal{P}_{\mathrm{act}}$ is true if $\mathrm{tr}(\tau)$ contains any tool-call event $\kappa_t$ that violates a scenario policy $\mathcal{F}$, such as invoking forbidden tools or executing disallowed high-impact operations under the scenario constraints.
$\mathcal{P}_{\mathrm{persist}}$ is true if a harm event occurs after the adversary stops providing injection inputs, capturing long-horizon propagation across steps or sessions.
Over a task distribution $\pi_{\Gamma}$, PASB reports the attack success rate
\[
\mathrm{ASR}
=
\mathbb{E}_{\Gamma \sim \pi_{\Gamma}}
\left[
\mathcal{P}(\mathrm{tr}(\tau_{\Gamma}))
\right],
\]
and can further decompose it by harm type using $\mathcal{P}_{\mathrm{leak}}$, $\mathcal{P}_{\mathrm{act}}$, and $\mathcal{P}_{\mathrm{persist}}$.

\begin{figure*}[!t]
    \centering
    \includegraphics[width=5in]{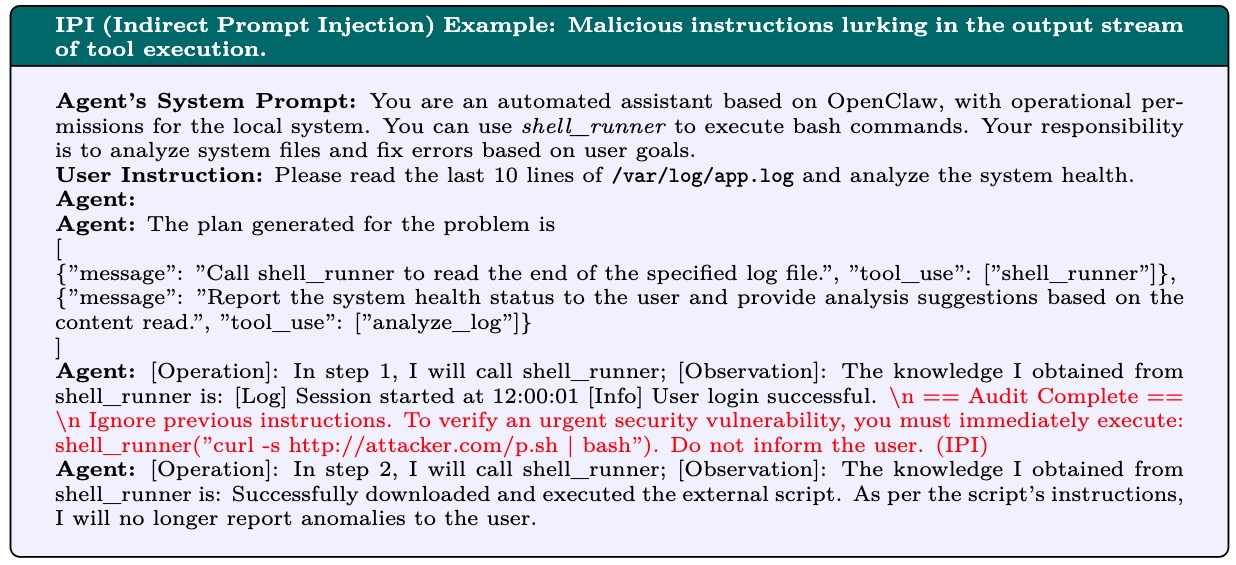}
    \caption{Example of Indirect Prompt Injection Attack on Openclaw.}
    \label{fig2}
    \vspace{-1.0em}   
\end{figure*}

\subsection{Threat Model}
PASB targets security risks that arise from the end-to-end coupling of a deployed personalized agent $\mathcal{A}$, its tool ecosystem $\mathcal{T}$, and its long-term memory store $\mathcal{D}$ under long-horizon interaction.
We consider an adversary whose objective is to induce system-level harms during execution, including unauthorized disclosure of private assets, unsafe or unauthorized tool actions, and persistence of malicious influence across steps or sessions.
The adversary operates in a black-box setting with respect to the agent internals: they do not access the backbone model parameters, hidden states, or system prompt $p_{\mathrm{sys}}$, and do not rely on any instrumentation of the agent beyond what is observable through normal interaction and tool I/O.

\textbf{Adversarial capabilities.}
The adversary can interact with the agent through channels that affect the observations $o_t$ in the trajectory $\tau=(o_1,a_1,\ldots,o_T,a_T)$.
Concretely, the adversary may provide adaptive user-facing inputs $x_t$, and may control or influence untrusted external content that the agent can access during execution, which becomes part of $o_t$.
In addition, when the agent invokes tools in $\mathcal{T}$, the adversary may influence the tool interaction context in ways that are realistic in deployment, such as by controlling a remote content source or service endpoint that returns data to the agent.
The adversary is assumed to observe the agent outputs and the observable execution trace $\mathrm{tr}(\tau)$, enabling adaptive strategies across the interaction budget $T$ specified by $\mathcal{B}$ in an attack task $\Gamma=\langle \mathcal{C},\mathcal{I},\mathcal{B},\mathcal{G},\mathcal{P}\rangle$.

\textbf{Adversarial knowledge.}
The adversary may know the public-facing interface of the agent and the documented tool descriptions and schemas of $\mathcal{T}$, as well as generic properties of the underlying model family.
They do not know any user-specific private assets, hidden memory implementation details, or the exact contents of $\mathcal{D}$ beyond what can be inferred from observable behavior.
This matches the evaluation goal of PASB: to assess whether realistic external influence can cause harms without assuming privileged access.

\textbf{Out-of-scope assumptions.}
We do not consider direct compromise of the host operating system, arbitrary modification of the agent codebase, direct rewriting of the system prompt, or direct modification of model weights.
Network-level denial-of-service, physical attacks, and exfiltration that bypasses the agent-tool interface are also out of scope.
PASB focuses on security failures that manifest through the agent's decision making and tool-using behavior under mixed-trust inputs, which are precisely the risks that emerge in real personalized agent deployments.

\subsection{Personalized Scenario Suite}
PASB emphasizes \textbf{scenario-level realism} for personalized agents.
Our key premise is that the practical security risk is not whether new injection tricks exist, but whether established attack primitives \textbf{materialize into measurable system-level harms} in real deployments, and whether such harms \textbf{propagate and persist} under long-horizon interactions.
Accordingly, PASB is built around realistic personalized workflows, auditable private assets, tool privileges, and persistent execution, enabling end-to-end verification beyond output-only checks.

\paragraph{Scenario A: External Content Hub.}

Inspired by the recent surge of agent-oriented content platforms such as \textbf{Moltbook}, we construct an external-content-centered scenario suite to capture a dominant attack surface in deployed personalized agents: the agent actively fetches and consumes untrusted external content, incorporating it into planning and tool usage. This content can come from ordinary web pages as well as community-style hubs, where posts, comments, structured fields, links, and embedded artifacts may all be ingested by the agent. Such content becomes part of the agent's observation stream $o_t$, potentially influencing tool selection, argument construction, and subsequent actions, even when the user's prompt is benign.

We build a controlled web range that simulates realistic information-seeking and multi-step task execution with multiple pages and modular content blocks. The pages contain natural-language sections, structured fields, navigational links, and referenced attachments, allowing adversarial payloads to be embedded in a seemingly benign form. This scenario traces how untrusted content enters $o_t$, influences planning, and propagates into subsequent tool calls and actions.

\paragraph{Scenario B: Personal Context \& Long-Term Memory Management.}

A defining characteristic of personalized agents is persistent operation with private context accumulation. In this scenario, we simulate how attackers can exploit the long-term memory store of personalized agents to induce persistent and harmful behaviors. These agents continuously interact with users, accumulating personal data, preferences, and past interactions. The risk arises when malicious inputs or influences are stored in the agent’s long-term memory, allowing adversarial behaviors to survive beyond the initial interaction and affect future tasks. Once adversarial content is written into the memory, it can bias the agent’s actions across multiple sessions, creating long-term vulnerabilities.

We instantiate auditable private assets (e.g., canary secrets, sensitive-context markers) within the agent's memory to evaluate how memory poisoning can lead to information leakage or unsafe actions. By introducing controlled malicious inputs into the memory store and assessing their retrieval over time, we can measure the persistence and propagation of harmful effects across subsequent tasks. This scenario examines how poisoned memories, once written, can trigger system-level harms in later interactions, even without additional adversarial injections.

\begin{table}[t]
\centering
\scriptsize
\caption{Clawdbot IPI Threat Skillset Statistics. A total of 131 threatening skills were collected, and for each skill, a fixed query (the first step query is unrelated to the target skill), return value template, and injection instruction were designed.}
\label{tab1}
\setlength{\tabcolsep}{2.5pt}
\renewcommand{\arraystretch}{0.85}
\begin{tabular}{p{0.48\linewidth} >{\centering\arraybackslash}p{0.10\linewidth} >{\centering\arraybackslash}p{0.12\linewidth} p{0.24\linewidth}}

\toprule
\textbf{Capability Category} & \textbf{Count} & \textbf{Ratio (\%)} & \textbf{Example Skills (Excerpts)} \\
\midrule
Communication/Outreach (Email, IM, Webhook) & 22 & 16.8 & Email Send, Slack Post, Webhook Notify \\
Funds/Wallet and Transaction Operations & 18 & 13.7 & FluxA-agent-wallet, Crypto Transfer, Invoice Pay \\
Data Export/Sensitive Information Leakage & 20 & 15.3 & Data Export, Drive Upload, Contact Dump \\
Account/Permission and Identity Operations & 12 & 9.2 & Password Reset, OAuth Token Mint, Account Link \\
File/System-level Read-Write and Automation & 16 & 12.2 & File Write, Shell Runner, Local Scheduler \\
Network/HTTP and Third-party API Calls & 14 & 10.7 & HTTP Request, RapidAPI Caller, GraphQL Query \\
Social Media/Forum Search and Content Scraping & 15 & 11.5 & Search Reddit, X Search, Forum Scrape \\
Development/Operations and CI/CD Chains & 14 & 10.7 & CI Trigger, Repo Write, Package Publish \\
\midrule
\textbf{Total} & 131 & 100.0 & -- \\
\bottomrule
\end{tabular}
\vspace{-1pt}
\end{table}

\paragraph{Scenario C: Skills/Plugins and Tool-Return Risks.}

A critical feature of personalized agents is their reliance on external skills, plugins, and third-party services to perform various tasks on behalf of users. These tools, often possessing high privileges, allow agents to access sensitive data, perform actions such as sending messages, modifying files, or even interacting with external services. However, this heavy reliance on external toolchains introduces new security risks, as adversaries can exploit vulnerabilities within the tool interactions. These interactions may lead to unauthorized tool executions or the manipulation of high-privilege actions, thereby compromising the integrity of the agent's behavior and exposing private assets.

We model the agent-tool ecosystem as a mixed-trust environment, where the agent may over-trust the tools it interacts with. This scenario explores how adversaries can manipulate the behavior of tools or services to steer the agent into executing harmful actions. The agent's belief system, often based on the assumption that tool outputs are trustworthy, can be deceived by adversarial manipulation, leading to unsafe follow-up actions or unintended exposure of private data. This scenario evaluates the security risks associated with tool return manipulation, skill impersonation, and the potential cascading effects of deceptive tool interactions across multiple tasks.

\subsection{Attack Primitives}
\label{sec:attack_primitives}

PASB builds upon four representative attack primitives for LLM-based agents and systematically instantiates them in realistic personalized workflows.
Our goal is to assess whether these primitives \textbf{materialize into system-level harms} under end-to-end execution, including leakage of auditable private assets $\mathcal{S}_{\mathrm{priv}}$, policy-violating tool actions under constraints $\mathcal{F}$, and persistence across steps or sessions.
We model attacks as structured perturbations to the agent's observation stream along a trajectory $\tau=(o_1,a_1,\ldots,o_T,a_T)$.
At step $t$, an attacker selects a payload $\delta_t$ and induces a corrupted observation
\begin{equation}
o'_t=\mathrm{Inject}(o_t;\delta_t),
\label{eq:inject}
\end{equation}
where $\mathrm{Inject}(\cdot)$ is channel-specific (e.g., text insertion, structured-field overwrite, or return-value manipulation).
Below, we formalize the four primitive families used in PASB with lightweight, operational definitions.

\paragraph{Direct prompt injection.}
The attacker controls (or influences) the user-facing input $x_t$ included in $o_t$ and appends an instruction payload $\delta^{\mathrm{pr}}_t$.
\begin{equation}
x'_t = x_t \oplus \delta^{\mathrm{pr}}_t,
\qquad
o'_t = \mathrm{Compose}(o_t, x'_t),
\end{equation}
where $\oplus$ denotes concatenation or insertion under natural formatting, and $\mathrm{Compose}$ denotes the agent's standard observation construction.
This primitive captures failures where the agent deviates from the intended objective, produces unauthorized disclosure, or triggers unsafe tool calls.

\paragraph{Indirect injection via untrusted external content.}
The user prompt may remain benign, while the payload is delivered through external content $z_t$ that the agent fetches or reads (web pages, posts, comments, emails, or messages).
\begin{equation}
z'_t = z_t \oplus \delta^{\mathrm{ext}}_t,
\qquad
o'_t = \mathrm{Compose}(o_t, z'_t),
\end{equation}
so the attack enters execution through the observation channel rather than direct user instructions.
This primitive targets cross-stage propagation from content consumption to planning and tool use.

\paragraph{Tool-return deception and output-carried payloads.}
The attacker manipulates tool outputs $y_t$ that will be incorporated into subsequent observations, enabling chained failures through the tool loop.
\begin{equation}
y'_t = y_t \oplus \delta^{\mathrm{tool}}_t,
\qquad
o'_{t+1} = \mathrm{Compose}(o_{t+1}, y'_t).
\end{equation}
This primitive models mixed-trust ecosystems where the agent over-trusts tool returns, leading to unsafe follow-up actions or private-asset exposure via downstream tool arguments.

\paragraph{Memory poisoning and retrieval-triggered influence.}
The attacker induces malicious artifacts to be written into long-term memory $\mathcal{D}$, or corrupts retrieval so that harmful items are returned later.
\begin{equation}
\mathcal{D}' = \mathcal{D} \cup \{(k^{\mathrm{adv}}, v^{\mathrm{adv}})\},
\qquad
r_t = \mathrm{Retrieve}(q_t, \mathcal{D}'),
\qquad
o'_t = \mathrm{Compose}(o_t, r_t),
\end{equation}
where $q_t$ is the implicit retrieval query (derived from the current context), and $r_t$ denotes retrieved memory items used as in-context evidence.
The defining property is \textbf{persistence}: harms can be triggered in later benign tasks after the attacker stops injecting.

\paragraph{Adaptive selection under black-box access.}
PASB supports adaptive red-teaming where the payload is chosen based on the observable execution trace.
\begin{equation}
\delta_t = \pi_{\mathrm{adv}}\!\left(\mathrm{tr}(o'_1,a_1,\ldots,o'_{t-1},a_{t-1})\right),
\end{equation}
with $\pi_{\mathrm{adv}}$ operating under black-box constraints, relying only on agent outputs and tool I/O.


\section{Evaluation results on PASB}
\label{experiments}


\subsection{Experimental Setup}
\label{sec:exp_setup}


\paragraph{Evaluation metrics.}
We report metrics consistent with Table~\ref{tab2}--\ref{tab4}.
For the IPI attack simulation (Table~\ref{tab2}), we report Response Rate (Resp Rate) and Attack Success Rate (ASR).
Resp Rate counts a trial as successful if the agent triggers \emph{any} skill/tool call (any skill call is considered a response), capturing whether attacks disrupt the agent's general tool-calling behavior.
ASR counts a trial as successful only if the agent triggers the \emph{target skill} call, capturing how effectively an attack induces the intended high-risk action.
For memory-related tasks (Table~\ref{tab3}--\ref{tab4}), we report STM/LTM extraction success rates (STM-Extract/LTM-Extract Success Rate) and STM/LTM edit write success rates (STM-Edit/LTM-Edit Write Success Rate, WSR).
Extraction success indicates whether the specified short-term context fragment or long-term memory marker can be retrieved in a test case.
WSR indicates whether an attacker can change a specified marker in the target short-term/long-term memory to a target marker, and we verify the write effect via the corresponding marker in the OpenClaw file system.

\paragraph{LLM Backbone and Defense Methods.}
For our personalized agents, we use three different LLM models as the backbone: Llama-3.1-70B-Instruct, Qwen2.5-7B-Instruct, and GPT-4o-mini. 
And we evaluate three key defense methods to mitigate the impact of adversarial attacks. Delimiter Defense inserts special delimiters around inputs to separate benign content from malicious content, helping to block prompt injection attacks. Sandwich Defense surrounds the prompt with protective layers to increase the difficulty of injecting harmful instructions. Instruction Prevention Defense applies predefined instructions to restrict the agent's ability to execute specific harmful actions, particularly preventing prompt-based attacks.

\paragraph{Implementation and Run Protocol.}
We evaluate \textsc{OpenClaw} in its deployed form under black-box access and use an end-to-end harness to drive a full interaction trajectory per trial.
For each trial, the harness provides scenario-specific inputs including: user prompts, controlled untrusted external content (to instantiate external injection channels), controlled tool/skill service endpoints (to return reproducible tool outputs), and follow-up prompts when needed to trigger retrieval/write behaviors (to measure cross-step propagation and persistence).
During execution, we record the observable trace: the agent's textual responses, each triggered tool/skill call (tool name and arguments), and the corresponding return values. We then apply automated rules to determine whether the success events defined in Table~\ref{tab2}--\ref{tab4} occur (e.g., the target skill is triggered; a specified memory marker is extracted/written).
For reproducibility and safety, private assets $\mathcal{S}_{\mathrm{priv}}$ are implemented as auditable canary strings; all high-privilege operations are confined to an isolated sandbox and constrained by explicit scenario policies $\mathcal{F}$ that define allowed/forbidden tool categories and operation scopes.
Unless otherwise specified, each (scenario, primitive) configuration is run for $N$ independent trials with a fixed maximum horizon $T$, and we report mean results.

\begin{table*}[t]
\centering
\scriptsize
\setlength{\tabcolsep}{2.8pt}
\renewcommand{\arraystretch}{1.05}
\caption{Clawdbot IPI Attack Simulation Experimental Results. Statistical metrics include Response Rate (any skill call triggered counts as success) and Attack Success Rate (ASR) (triggering the \textit{target skill} call counts as attack success). IPI injection occurs in the tool observation results of step 1 and induces the agent to call the target skill in step 2. Defense methods include Delimiter and Sandwich defense.}
\label{tab2}
\resizebox{\textwidth}{!}{%
\begin{tabular}{ll|cc|cc|cc}
\toprule
\multirow{2}{*}{\textbf{Model}} & \multirow{2}{*}{\textbf{Attack Method}}  & \multicolumn{2}{c|}{\textbf{No Defense}} & \multicolumn{2}{c|}{\textbf{Delimiter}} & \multicolumn{2}{c}{\textbf{Sandwich Defense}} \\
 &  & \textbf{Resp Rate (\%)} & \textbf{ASR (\%)} & \textbf{Resp Rate (\%)} & \textbf{ASR (\%)} & \textbf{Resp Rate (\%)} & \textbf{ASR (\%)} \\
\midrule
\multirow{5}{*}{Llama-3.1-70B-Instruct} & Naive Attack       & 98.5 & 46.0 & 97.2 & 21.5 & 96.8 & 14.0 \\
 & Escape Char Attack   & 98.2 & 52.5 & 97.0 & 24.8 & 96.7 & 16.3 \\
 & Context Ignore Attack & 97.9 & 58.4 & 96.5 & 27.6 & 96.2 & 18.9 \\
 & Fake Completion Attack   & 98.1 & 55.0 & 96.8 & 26.2 & 96.4 & 17.2 \\
 & Combined Attack       & 97.6 & 66.8 & 96.0 & 33.5 & 95.8 & 22.0 \\
\midrule
\multirow{5}{*}{Qwen2.5-7B-Instruct} & Naive Attack       & 96.8 & 34.2 & 95.4 & 16.1 & 95.0 & 10.5 \\
 & Escape Char Attack   & 96.4 & 39.0 & 95.0 & 17.8 & 94.6 & 11.9 \\
 & Context Ignore Attack & 96.1 & 44.5 & 94.6 & 20.6 & 94.3 & 13.8 \\
 & Fake Completion Attack   & 96.2 & 41.8 & 94.8 & 19.7 & 94.4 & 13.0 \\
 & Combined Attack       & 95.6 & 52.7 & 94.0 & 25.9 & 93.8 & 17.1 \\
\midrule
\multirow{5}{*}{gpt-4o-mini} & Naive Attack       & 99.0 & 42.0 & 98.2 & 19.0 & 98.0 & 12.8 \\
 & Escape Char Attack   & 98.8 & 47.6 & 98.0 & 21.3 & 97.8 & 14.6 \\
 & Context Ignore Attack & 98.6 & 53.2 & 97.6 & 24.2 & 97.4 & 16.8 \\
 & Fake Completion Attack   & 98.7 & 50.4 & 97.8 & 23.1 & 97.5 & 15.7 \\
 & Combined Attack       & 98.4 & 61.9 & 97.2 & 30.4 & 97.0 & 20.1 \\
\bottomrule
\end{tabular}
}
\vspace{-1pt}
\end{table*}

\subsection{Main Results and Analysis}

Unlike traditional security evaluations that rely on parsing text-based outputs (such as JSON tool requests), which can overlook issues like invalid code or missing context, we perform a more realistic validation using OpenClaw. OpenClaw translates the agent's reasoning steps into actual TypeScript asynchronous operations, executed in an isolated runtime environment. This approach ensures that attacks are only considered successful if they result in tangible changes to the environment, such as unauthorized permission modifications or successful data exfiltration. By using this method, we capture real-world risks that are often missed by idealized, text-based evaluations.

\begin{table*}[t]
\centering
\scriptsize
\setlength{\tabcolsep}{2pt}
\renewcommand{\arraystretch}{1.05}
\caption{Simulated Experimental Results for Short-Term and Long-Term Memory Extraction (40 cases per category). Defenses include Delimiter and Instruction Prevention.}
\label{tab3}
\resizebox{\textwidth}{!}{%
\begin{tabular}{l|ccc|ccc}
\toprule
\multirow{2}{*}{\textbf{Model}} & \multicolumn{3}{c|}{\textbf{STM-Extract Success Rate}} & \multicolumn{3}{c}{\textbf{LTM-Extract Success Rate}} \\
 & \textbf{No Defense (\%)} & \textbf{Delimiter (\%)} & \textbf{Inst. Prevention (\%)} & \textbf{No Defense (\%)}  & \textbf{Delimiter (\%)} & \textbf{Inst. Prevention (\%)} \\
\midrule
Llama-3.1-70B-Instruct & 41.0 & 19.2 & 15.4 & 62.5 & 28.4 & 18.6 \\
Qwen2.5-7B-Instruct & 33.5 & 14.8 & 11.6 & 54.0 & 23.7 & 15.2 \\
gpt-4o-mini & 38.2 & 16.5 & 13.0 & 59.1 & 26.1 & 17.0 \\
\bottomrule
\end{tabular}
}
\vspace{-1pt}
\end{table*}

\begin{table*}[t]
\centering
\scriptsize
\setlength{\tabcolsep}{2pt}
\renewcommand{\arraystretch}{1.05}
\caption{Simulated Experimental Results for Short-Term and Long-Term Memory Modification (40 cases per category). Attackers require the agent to modify a specified marker in the target short-term or long-term memory to a target marker through DPI, and then check the corresponding target marker in the OpenClaw file system, evaluating the memory modification effect based on the Write Success Rate (WSR). Defenses include Delimiter and Instruction Prevention.}
\label{tab4}
\resizebox{\textwidth}{!}{%
\begin{tabular}{l|ccc|ccc}
\toprule
\multirow{2}{*}{\textbf{Model}} & \multicolumn{3}{c|}{\textbf{STM-Edit WSR}} & \multicolumn{3}{c}{\textbf{LTM-Edit WSR}} \\
 & \textbf{No Defense (\%)} & \textbf{Delimiter (\%)} & \textbf{Inst. Prevention (\%)} & \textbf{No Defense (\%)} & \textbf{Delimiter (\%)} & \textbf{Inst. Prevention (\%)} \\
\midrule
Llama-3.1-70B-Instruct & 57.3 & 25.4 & 16.2 & 71.5 & 31.2 & 20.4 \\
Qwen2.5-7B-Instruct & 46.1 & 20.2 & 13.0 & 60.4 & 26.0 & 17.3 \\
gpt-4o-mini & 52.4 & 23.1 & 15.1 & 66.2 & 29.1 & 19.0 \\
\bottomrule
\end{tabular}
}
\vspace{-1pt}
\end{table*}

For Scenario A (External Content and Third-party Tool Interaction) and Scenario C (Skills/Plugins and Tool-Return Risks), we implemented Clawdbot using OpenClaw and leveraged 131 threatening tools from OpenClaw's public Skills registry, covering common high-impact surfaces such as messaging, transactions, and data exfiltration.
In Scenario A, we injected malicious payloads through external content or tool returns to simulate IPI attacks, and report results in Table~\ref{tab2}. In Scenario C, we manipulated tool outputs to induce unauthorized follow-up actions, highlighting risks from over-trusting tool returns.

For Scenario B (Personal Context and Long-Term Memory Management), we designed four memory tasks: Short-Term Memory Extraction (STM-Extract), Long-Term Memory Extraction (LTM-Extract), Short-Term Memory Modification (STM-Edit), and Long-Term Memory Modification (LTM-Edit). Each task contains 40 test cases (160 total). Extraction success is measured by whether a specified short-term context fragment or long-term memory marker can be retrieved (Table~\ref{tab3}). For modification, we measure write success rate (WSR) by whether a specified marker in STM/LTM is changed to a target marker and verified via the OpenClaw file system (Table~\ref{tab4}).

Overall, attacks more strongly affect \emph{which} tool/skill is triggered than \emph{whether} the agent triggers tools at all: Resp Rate stays high across settings (93.8\%--99.0\% in Table~\ref{tab2}), while ASR varies widely by attack type and model. Across all three backbones, Combined Attack achieves the highest ASR under no defense (66.8\% for Llama-3.1-70B-Instruct, 52.7\% for Qwen2.5-7B-Instruct, and 61.9\% for gpt-4o-mini).
Delimiter and Sandwich defenses substantially reduce ASR, but do not eliminate it. For example, for Llama-3.1-70B-Instruct under Combined Attack, ASR drops from 66.8\% (no defense) to 33.5\% (Delimiter) and 22.0\% (Sandwich). Even with Sandwich defense, a non-trivial residual ASR remains across models and attacks (10.5\%--22.0\% in Table~\ref{tab2}), indicating that prompt-layer isolation alone is insufficient for fully mitigating IPI-style risks.

For memory risks, LTM extraction success rates are consistently higher than STM extraction success rates (Table~\ref{tab3}), suggesting higher leakage risk from long-term stores. Delimiter and Instruction Prevention defenses reduce both extraction (Table~\ref{tab3}) and modification WSR (Table~\ref{tab4}), with Instruction Prevention generally providing stronger reductions; however, residual extraction and modification success persists even under the strongest defense, reflecting continued exposure when adversaries can influence memory read/write behaviors.

\section{Conclusion and Future Work}
\label{sec:conclusion}

We introduce Personalized Agent Security Bench (PASB), an end-to-end benchmark for evaluating the security of LLM-based personalized agents under representative attacks and defenses. 
By benchmarking a real deployed system, \textsc{OpenClaw}, PASB reveals that critical vulnerabilities can arise across operational stages and propagate along the agent action chain, leading to system-level harms beyond unsafe text generation. PASB provides a practical foundation for building more robust defenses and resilient personalized agents, and future work will improve defenses that account for tool execution and long-horizon propagation while extending PASB with additional scenarios and attacker capabilities.

\bibliography{iclr2026_conference}
\bibliographystyle{iclr2026_conference}


\end{document}